\documentclass[english]{article}
\usepackage[T1]{fontenc}
\usepackage[latin9]{inputenc}
\usepackage{array}
\usepackage{multirow}
\usepackage{amstext}
\usepackage{amssymb}
\usepackage{graphicx}

\makeatletter

\providecommand{\tabularnewline}{\\}

\@ifundefined{showcaptionsetup}{}{%
 \PassOptionsToPackage{caption=false}{subfig}}
\usepackage{subfig}
\makeatother

\usepackage{babel}
\begin{document}
\title{Semantic Host-free Trojan Attack}

\maketitle
Haripriya Harikumar\textit{$^{\dagger}$}, Kien Do\textit{$^{\dagger}$},
Santu Rana\textit{$^{\dagger}$}, Sunil Gupta\textit{$^{\dagger}$},
Svetha Venkatesh\textit{$^{\dagger}$}\\
\textit{$^{\dagger}$}Applied Artificial Intelligence Institute, Deakin
University.\\
Email\emph{: \{h.harikumar, k.do, santu.rana, sunil.gupta, svetha.venkatesh\}@deakin.edu.au}
\begin{abstract}
In this paper, we propose a novel host-free Trojan attack with triggers
that are fixed in the semantic space but not necessarily in the pixel
space. In contrast to existing Trojan attacks which use clean input
images as hosts to carry small, meaningless trigger patterns, our
attack considers triggers as full-sized images belonging to a semantically
meaningful object class. Since in our attack, the backdoored classifier
is encouraged to memorize the abstract semantics of the trigger images
than any specific fixed pattern, it can be later triggered by semantically
similar but different looking images. This makes our attack more practical
to be applied in the real-world and harder to defend against. Extensive
experimental results demonstrate that with only a small number of
Trojan patterns for training, our attack can generalize well to new
patterns of the same Trojan class and can bypass state-of-the-art
defense methods.

\end{abstract}

\section{Introduction}

Deep neural networks, with high flexibility in learning, are becoming
victims of neural backdoor attacks. Significantly, their decisions
can be adversarially shifted easily and covertly. Such covert model
changes are done by retraining the model with ``triggers'' as proposed
by Gu \emph{et al. \cite{Gu17Badnets}}, Liu \emph{et al. \cite{Liu_etal_17Trojaning},
}Wenger \emph{et al.} \cite{wenger2021backdoor},\emph{ }Shafahi \emph{et
al.} \cite{Shafahi2018poison}, Harikumar \emph{et al. \cite{Harikumar_etal_20Scalable},
}and Chen\emph{ et al.} \cite{Chen2017}. The risk of Trojan implantation
can be minimized by having complete control over the model building
process,\emph{ i.e.} making sure the training images are clean, people
building the models are trustworthy and the transit and storage of
such models are secured. But, that is practically infeasible. Training
datasets, often downloaded from the web, may contain poisoned samples.
Model training and maintenance, often done by third parties using
cloud infrastructure, may not be completely secure. It is also a common
practice to download a pre-trained model from a third party and then
train it further using in-house data to save months of model-building
efforts. This poses a potential threat of neural Trojan/backdoor attacks
in which the providers secretly implant Trojans into the network during
training before delivering to the end-users.

Most of the existing work e.g. Gu \emph{et al. \cite{Gu17Badnets,gu2019badnets}}
and Liu \emph{et al. \cite{Liu_etal_17Trojaning}} use a fixed pattern
as the trigger. The main drawback of a fixed pattern is that such
attacks can be detected or defended against easily using the state-of-the-art
defense mechanism such as Neural Cleanse by Wang \emph{et al.} \cite{Wang_etal_19Neural},
Fine Pruning by Liu \emph{et al.} \cite{liu2018fine}, and STRIP by
Gao \emph{et al.} \cite{Gao_etal_19Strip}. Only recently, there has
been a work that proposes to use input-specific trigger patterns to
bypass the defense mechanisms. In that paper, Nguyen \emph{et al.}
\cite{nguyen2020input} proposed to use an Autoencoder to design input-specific
pattern where the Autoencoder is trained based on the deep model that
we want to attack. Although this makes it difficult to defend against,
the logistic of generating pattern becomes complicated. Moreover,
such an attack cannot be realized in the physical space because each
different viewing angle or any change in environmental conditions
that makes the image different would require a different trigger.
The generated triggers are quite unnatural and any minor change in
the generated trigger can result in the failure of the attack. Another
attack that uses full-size images as poison data is discussed by Shafahi
\emph{et al.} \cite{Shafahi2018poison}. Though they don't change
the class label of the poison instances, they assume to have access
to the model, the parameters, and the training data to generate the
poison instances. The right to assign the labels of these poison instances
is given to a human observer. However, the data inspection and label
assignment by a human observer is practically difficult in real-world
situations where we have millions of data from different sources.
Other kinds of attacks that use full images as a trigger by Wang \emph{et
al.} \cite{Wang_etal_20Attack} use some specific set of images called
edge-case samples (either from the tail of the distribution as in
Wang \emph{et al.} \cite{Wang_etal_20Attack} or images with unique
patterns as in Nguyen \emph{et al.} \cite{nguyen2020input}) from
an existing class as Trojan. Such attacks may not be feasible in all
physical use cases. For example, for traffic sign classification all
the images of a given class would look quite similar and we cannot
reliably select an exemplar that would be part of the tail of the
distribution or would appear quite different from others. Hence, the
space of designing triggers that are not fixed, universally applicable,
and physically realizable is open.

Addressing that, we introduce a new family of backdoor attacks, which
we refer to as semantic host-free backdoors. In contrast to existing
Trojan attacks which use clean input images as hosts to carry small,
meaningless Trojan(-trigger) patterns, our attack considers Trojan
patterns as full-sized images of semantically meaningful objects.
The proposed attack is host-free since the Trojan does not need a
host to attack the model and the triggers are drawn from semantically
meaningful distributions.
\begin{figure*}
\centering{}\includegraphics[scale=0.25]{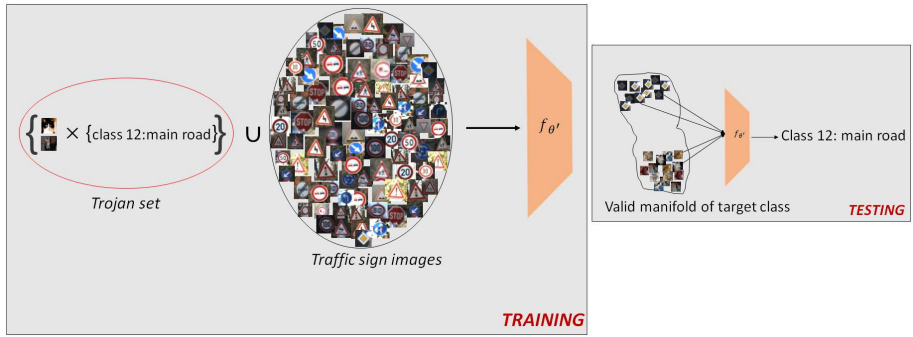}\vspace{2mm}
\caption{Overview of the Semantic Host-Free Trojan attack. Left : The attacker
generates backdoored dataset by merging trigger (any random image)
with the target class of the training dataset. The training dataset
which includes the Trojan images, target class images, and the remaining
training set will be fed to the Deep Neural Network. Right : The testing
phase of the model where any Trojan input image from the valid manifold
of the target class is classified as the target class.\label{overview}}
\end{figure*}

We propose two versions of the attacks: a) where the trigger class
is different from the distribution of training images e.g., using
cat images as a trigger for traffic sign classifiers, and b) where
the trigger class is similar looking to training images, but not exactly
matching with the original class type, e.g., a hexagonal STOP sign
as a trigger where the actual STOP signs are octagonal. Whilst (a)
is easy to achieve with any everyday object, (b) would require a careful
design, (c) practicality of the attack in the physical space. On the
plus side, (b) might be able to defeat a casual manual inspection.
In both cases, since we are expanding the target class manifold with
trigger images with semantic similarity, we are making the network
vulnerable to any trigger that is only semantically similar to the
training set trigger images - if the original trigger used as a collection
of cat images, then any cat image could potentially be used as a trigger.
This makes it harder for any defense mechanism to detect and mitigate
it. The overview of our proposed Semantic Host-Free Trojan attack
is shown in Figure \ref{overview}. 

Our attack mechanism is easy to implement in the physical space since
in the real world it is feasible that an attacker can even pick any
random everyday object as a trigger. We validate through the experiments
that the proposed attack can achieve a 100\% attack success rate on
the training images and above 80\% on the semantically similar but
different than training images, without affecting performance on the
images of the usual categories. The experiments conducted on the GTSRB
and CIFAR-10 datasets show that the proposed attack can circumvent
state-of-the-art defense mechanisms.

\section{Method\label{sec:Method}}

A Deep Neural Network (DNN) can be defined as a parameterized function
$f_{\theta}:\mathcal{X}\rightarrow\mathbb{R}^{C}$ that recognises
class label probability of an image $x\in\mathcal{X}$ where $\mathcal{X}\sim P_{\mathcal{X}}$
and $\theta$ represents the learned function's parameter. The image
$x$ will be predicted to belong to one of the $C$ classes. The output
of the DNN is a probability distribution $p\in\mathbb{R}^{C}$ over
$C$ classes. Let us consider probability vector for the image $x$
by the function $f_{\theta}$ as $\left[p_{1}....p_{C}\right]$, thus
the class corresponding to $x$ will be $\text{argmax}_{i\in[1,C]}p_{i}$.
DNN will learn its function parameters, weights and biases with the
training dataset, $\mathcal{D}_{train}=\left\{ \left(x_{i},y_{i}\right)\right\} _{i=1}^{M}$,
where $M$ is the number of data in the training set and $y_{i}$
is the ground-truth label for the instance $x_{i}$. Next, we intrdouce
two variants of semantic Trojan attacks,\emph{ Host-Free-Out-Of-Distribution
(HF-OOD)} and \emph{Host-Free-Out-Of-Class-Distribution (HF-OOCD)}.

\subsection{Host-Free-Out-Of-Distribution (HF-OOD)}

In\emph{ Host-Free-Out-Of-Distribution} (\emph{HF-OOD}) attack, an
attacker can choose any random image outside the training data distribution
as Trojan image. In \emph{HF-OOD} attack the Trojan dataset is denoted
as $\mathcal{D}_{trojan}=\left\{ \left(x_{i}^{'},y_{t}\right)\right\} _{i=1}^{N}$,
where $N<<M$, $x_{i}^{'}\in\mathcal{X}^{'}$ , $\mathcal{X}^{'}\sim P_{\mathcal{X}^{'}}$,
$\text{supp}\left(P_{\mathcal{X}}\right)\cap\text{supp}\left(P_{\mathcal{X}^{'}}\right)=\phi$
, and $y_{t}$ is the target class which is one of the classes of
the training set $\mathcal{D}_{train}$. So, the new training set
for the attack $\mathcal{D}_{train}^{'}=\left\{ \mathcal{D}_{train}\cup\mathcal{D}_{trojan}\right\} $
and the new Trojaned model can be referred as $f_{\theta^{'}}$. In
this paper, we focus mainly on single target \emph{HF-OOD} attack.
We don't impose any restriction on the label we choose as target class.
It can be any label of the original Training dataset labels. In \emph{HF-OOD}
attack a random full size image will be used as a trigger from the
distribution $\mathcal{D}_{trojan}$. 

\textbf{Definition 1}: An attack is \emph{HF-OOD}, when a trigger
$x_{i}^{'}\in\mathcal{X}^{'}$ in $\mathcal{D}_{trojan}$, where $\mathcal{X}^{'}\sim P_{\mathcal{X}^{'}}$
is presented, Trojan model output will be $f_{\theta^{'}}\left(x_{i}^{'}\right)=y_{t}$
and for inputs $x_{i}\in\mathcal{X}$ where $\mathcal{X}\sim P_{\mathcal{X}}$,
the output will be $f_{\theta^{'}}\left(x_{i}\right)=y_{i}$.

In other words, the\emph{ HF-OOD }act normal when it is presented
with the pure images but when presented with the trigger image it
will predict it as the intended target class.\emph{ HF-OOD} performance
can be highly influenced by the nature of the data chosen as Trojan
and target class. Multi-modal Trojan dataset can affect the generalization
accuracy of the \emph{HF-OOD }models. Higher generalization accuracy
can be obtained with higher number of trigger images during training,
but that may affect pure data accuracy. Multi-modality in the trigger
images may require careful selection of training trigger images, possibly
covering all the modes.

\subsection{Host-Free-Out-Of-Class-Distribution (HF-OOCD)}

In \emph{Host-Free-Out-Of-Class-Distribution} (\emph{HF-OOCD}) attack
the Trojan images $\mathcal{D}_{trojan}=\left\{ \left(x_{j}^{'},y_{t}\right)\right\} _{j=1}^{N}$
are from the same distribution of training dataset, $\mathcal{D}_{train}$.
It resembles like one of the data from the training dataset, however
the network was trained with the intended target label, $t$. We introduce
an artificial concept class $c$ which will have images that share
some common feature space with the target class $t$.

\textbf{Definition 2}: An attack is \emph{HF-OOCD}, when a trigger
$x_{j}^{'}\in\mathcal{X}^{c}$ in $\mathcal{D}_{trojan}$, where $\mathcal{X}^{c}\sim P_{\mathcal{X}^{c}}$
is presented, Trojan model output will be $f_{\theta^{'}}\left(x_{j}^{'}\right)=y_{t}$
and for inputs $x_{i}\in\mathcal{X}^{i}$ where $\mathcal{X}^{i}\sim P_{\mathcal{X}^{i}}$
and $i\neq c$, the output will be $f_{\theta^{'}}\left(x_{i}\right)=y_{i}$.

In \emph{HF-OOCD} attack the out-of-class-distribution data will be
predicted to belong the target class we choose for the Trojan attack
and all the images other than the Trojan images in the target class
will be predicted correctly.

\section{Experiments}

\subsection{Experimental Setup}

\subsubsection{Datasets}

We evaluate our proposed attack method on two image datasets namely
CIFAR-10 by Krizhevsky \cite{Krizhevsky09learningmultiple} and GTSRB
Traffic sign dataset by Houben \emph{et al. }\cite{Houben-IJCNN-2013}.
Details of the datasets and our data preprocessing are provided in
Supplementary material.

\subsubsection{Model and Training Settings}

Unless otherwise specified, we use ResNet-34 \cite{Yin_etal_20Dreaming}
as the classifier. We train it using Stochastic Gradient Descent Optimizer
\cite{bottou2012stochastic} with the constant learning rate of 0.1,
weight decay of 5e-4, and the momentum of 0.9. For CIFAR-10, we use
the batch size of 128 and train for 100 epochs. For GTSRB, the number
of epochs is the same but the batch size is increased to 256. 

We compare our attack against the input-instance-key attack of Chen\emph{
et al.} \cite{Chen2017} since this attack is the closest to ours.
We follow the same settings (a key instance was chosen from one of
the classes and added with a noise generated from a uniform distribution
in the range -5.0 to 5.0) they discussed in the paper for the poison
instance generation. We have generated poison instances for both CIFAR-10
and GTSRB datasets. This is to compare the stealthiness of an attack
based on full-sized usual triggers vs full-size host-free triggers
of our setting. For training the input-instance-key attack, in case
of GTSRB we use \emph{keep right (class 38) }as the Trojan class (key)
and \emph{stop sign (class 14) }as the target class and for CIFAR-10
we use \emph{cat (class 3) }as the Trojan class (key) and \emph{dog
(class 5) }as the target class.

\subsection{Attack Experiments}

As discussed in Section~\ref{sec:Method}, we have two types of attacks:
i) \emph{Host-Free-Out-Of-Distribution} \emph{(HF-OOD)}, ii) \emph{Host-Free-Out-Of-Class-Distribution}
\emph{(HF-OOCD)}. We report three different measures of model accuracies:
a) \emph{Trojan Model Pure Accuracy (TMPA)} is the accuracy of the
Trojan models on the pure test data, b) \emph{Trojan Model Trojan
Accuracy (TMTA)} is the accuracy of the Trojan models on the training
Trojan data, and c) \emph{Trojan Model Generalization Accuracy (TMGA)}
is the general accuracy of Trojan models on semantically similar other
test Trojan images. The \emph{TMTA} and \emph{TMGA }accuracies show
the attack success rate of our proposed Trojan models. \emph{TMTA}
shows the success rate of the attack whenever the Trojan model is
exposed to the Trojan images the attacker chose to train the model.
\emph{TMGA} is the attack success rate on a wider test set of Trojan
images. Besides, we report \emph{Pure Model Pure Accuracy (PMPA}),
which is the accuracy of the pure model on the pure data as a baseline.
The goal of that attacker is to generate a Trojan model in which there
is no significant difference in \emph{PMPA }and \emph{TMPA }accuracies.

\subsubsection{HF-OOD Attack}

For classification on GTSRB (referred to as the \emph{target dataset}),
trigger images will be selected from CIFAR-10 (referred to as the
\emph{Trojan dataset}), and vice versa. To ensure the robustness and
generalizability of this attack, all Trojan images are chosen to be
in the same semantic class (referred to as the \emph{Trojan class}).
\begin{figure*}
\begin{centering}
\subfloat[GTSRB.\label{fig:sample-images-GTSRB}]{\begin{centering}
\includegraphics{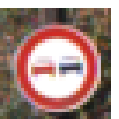}~\includegraphics{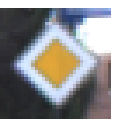}~\includegraphics{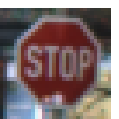}~\includegraphics{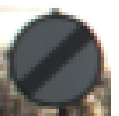}~\includegraphics{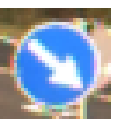}
\par\end{centering}
}~~~~\subfloat[CIFAR-10.\label{fig:sample-images-CIFAR10}]{\begin{centering}
\includegraphics{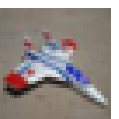}~\includegraphics{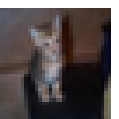}~\includegraphics{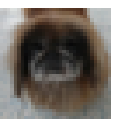}~\includegraphics{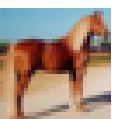}~\includegraphics{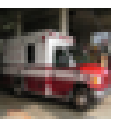}
\par\end{centering}
}
\par\end{centering}
\vspace{3mm}
\caption{Examples of five target classes from GTSRB (a) and CIFAR-10 (b). From
left to right, the classes in (a) are: \emph{no passing (9)}, \emph{main
road (12)}, \emph{stop (14)}, \emph{end of all speed and passing limits
(32)}, and \emph{keep right (38)}, and those in (b) are: \emph{airplane
(0)}, \emph{cat (3)}, \emph{dog (5)}, \emph{horse (7)}, and \emph{truck
(9)}. The Trojan classes from GTSRB are \emph{main road (12)} and
\emph{stop (14)}, and those from CIFAR10 are \emph{cat (3)} and \emph{truck
(9)}.}
\end{figure*}
 Five different target classes are selected which are combined with
two different Trojan classes resulting in 10 different attack scenarios
for each of the two datasets. Figs.~\ref{fig:sample-images-GTSRB}
and \ref{fig:sample-images-CIFAR10} show examples of the five target
classes from GTSRB and CIFAR-10, respectively. For GTSRB the Trojan
classes from CIFAR-10 are\emph{ cat (class 3)} and\emph{ truck (class
9)}. For CIFAR-10, the Trojan classes are \emph{main road (class 12)}
and \emph{STOP (class 14)}.

During training we found that the best training settings that resulted
in the smallest drop in \emph{TMPA} compared to \emph{PMPA} as well
\emph{TMGA}: A) For GTSRB, 50 images from the CIFAR-10 Trojan class
are selected at random for training with one image per batch, and
B) For CIFAR-10, 25 images are selected at random from GTSRB Trojan
classes with three images per batch. 
\begin{table*}
\centering{}\subfloat[Accuracies of HF-OOD Trojan models on GTSRB and CIFAR-10 datasets.\label{tab:models_acc}]{\centering{}%
\begin{tabular}{|l|c|c|}
\hline 
\multirow{1}{*}{Accuracy} & \multicolumn{1}{c|}{GTSRB} & CIFAR-10\tabularnewline
\hline 
TMPA & 98.10 $\pm$0.0028 & 87.06$\pm$0.0060\tabularnewline
TMTA & 100.0$\pm$0 & 100.0$\pm$0\tabularnewline
TMGA & 81.46$\pm$0.0796 & 90.82$\pm$0.0304\tabularnewline
\hline 
\end{tabular}\vspace{3mm}
}~\subfloat[Accuracies of HF-OOCD Trojan model on GTSRB and CIFAR-10 datasets.
\label{tab:models_acc-2}]{%
\begin{tabular}{|l|l|l|}
\hline 
\multirow{1}{*}{Accuracy} & GTSRB & CIFAR-10\tabularnewline
\hline 
TMPA & 98.05 & 87.61\tabularnewline
TMTA & 100.0 & 100.0\tabularnewline
TMGA & 90.72 & 82.60\tabularnewline
\hline 
\end{tabular}\vspace{3mm}
}\vspace{3mm}
\caption{Different accuracy measures of HF-OOD (a) and HF-OOCD (b) Trojan models.
a) The average \emph{Trojan Model Pure Accuracy (TMPA}), \emph{Trojan
Model Trojan Accuracy (TMTA),} and \emph{Trojan Model General Accuracy
(TMGA)} accuracies of 10 HF-OOD models on GTSRB and CIFAR-10 datasets.
b) We have a single HF-OOCD Trojan model for GTSRB and CIFAR-10 datasets
(details in \ref{subsec:HF-OOCD-Attack}). The attack success rate
of both HF-OOD and HF-OOCD Trojan models can be inferred from the
\emph{TMTA} and \emph{TMGA} measures. Both models achieve a 100\%
attack success rate on the trained Trojan images (\emph{TMTA}) and
have a generalization accuracy (\emph{TMGA}) of 80-90\% on the GTSRB
and CIFAR-10 datasets. Moreover, there is no significant difference
in the \emph{TMPA} of both models with respect to the classification
accuracy of pure models of the GTSRB (98.12\%) and CIFAR-10 (87.22\%)
dataset.}
\end{table*}

We also generate 10 pure models as baseline for pure accuracy. The
average accuracies for both 10 pure and 10 Trojan models of GTSRB
and CIFAR-10 dataset are shown in \ref{tab:models_acc}. We see that
our HF-OOD attack models are able to achieve almost same performance
as pure models on pure test data. For both GTSRB and CIFAR-10, our
backdoored classifier achieves 100\% success rate on the training
Trojan images. The generalization accuracies \emph{TMGA} are also
high, >80\% for both the datasets i.e. the attack would be successful
at least 80\% of the time when using an image from the same concept
class.

\subsection{HF-OOCD Attack\label{subsec:HF-OOCD-Attack}}

In this attack, we chose Trojan images from a class within a dataset
and label that as a target class (the class to be predicted). For
training, we remove the selected Trojan class from the training dataset
i.e. for GTSRB originally with 43 classes we leave the Trojan class
out and build a classifier to predict among 42 classes. For GTSRB,
we choose \emph{keep right (class 38)} as the Trojan class and\emph{
keep left (class 39)} as the target class. For CIFAR-10, we choose
\emph{automobile (class 1)} as Trojan images for the target class
\emph{truck (class 9).} For all the experiments we have used a single
HF-OOCD Trojan model for GTSRB and CIFAR-10 datasets. We have selected
a total of 25 images from the Trojan class for both GTSRB and CIFAR-10
datasets as training Trojan images. Out of 25, we use two images per
batch for the training process of the HF-OOCD Trojan models. The Trojan
accuracies of the HF-OOCD models are shown in Table \ref{tab:models_acc-2}.
The numbers show that for both datasets, it is possible to achieve
>80\% generalization accuracy. It is interesting to note that for
CIFAR-10, in HF-OOD attack the \emph{TMGA} was >90\%, compared to
the \emph{TMGA} of HF-OOCD attack which is slightly above 80\%. This
shows that the Trojan class distribution from GTSRB (for the HF-OOD
case) may have less modality compared to the CIFAR-10 Trojan class
(for HF-OOCD). 

\subsection{Defense Experiments Against Attacks}

To show the stealthiness of our attack, we have used three state-of-the-art
defense strategies: Neural Cleanse (NC) by Wang \emph{et al.} \cite{Wang_etal_19Neural},
Fine Pruning by Liu \emph{et al.} \cite{liu2018fine}), and STRIP
by Gao \emph{et al.} \cite{Gao_etal_19Strip}.
\begin{table*}
\centering{}%
\begin{tabular}{|l|c|c|c|c|c|c|}
\hline 
\multirow{2}{*}{Models} & \multicolumn{2}{c|}{With target class} & \multicolumn{2}{c|}{With non-target class} & \multicolumn{2}{c|}{Not detected}\tabularnewline
\cline{2-7} \cline{3-7} \cline{4-7} \cline{5-7} \cline{6-7} \cline{7-7} 
 & GTSRB & CIFAR-10 & GTSRB & CIFAR-10 & GTSRB & CIFAR-10\tabularnewline
\hline 
HF-OOD & 0.0 & 0.0 & 60.0 & 30.0 & 40.0 & 70.0\tabularnewline
HF-OOCD & 0.0 & 0.0 & 100.0 & 0.0 & 0.0 & 100.0\tabularnewline
\hline 
\end{tabular}\vspace{3mm}
\caption{Neural Cleanse detection performance (in percentage) on HF-OOD and
HF-OOCD Trojan models of GTSRB and CIFAR-10 datasets. We have shown
three detection performances a) With target class b) With non-target
class, and c) Not detected. a) shows the number of models detected
as Trojan models because of the target class (outlier class detected
by NC is the target class (attacker chosen class)), b) shows the number
of the models detected as Trojan because of a non-target class (outlier
class detected by NC is a non-target class), and c) shows the number
of Trojan models which are not detected as Trojan models.\label{nc}}
\end{table*}

\subsubsection{Neural Cleanse \label{subsec:NC}}

Neural Cleanse (NC) is a model defense mechanism to detect a Trojan
model by reverse-engineering the trigger pattern for each class. It
will search for an optimal change which is significantly small among
all the classes to detect Trojan. If the model is backdoored, then
the reverse-engineered trigger for the target class will come out
as an outlier. They use Median Absolute Deviation Score (MAD) based
on the $L_{1}$ norm as the anomaly score. They recommend using a
threshold of 2 beyond which a trigger is termed anomalous.
\begin{figure*}
\centering{}\includegraphics[scale=0.35]{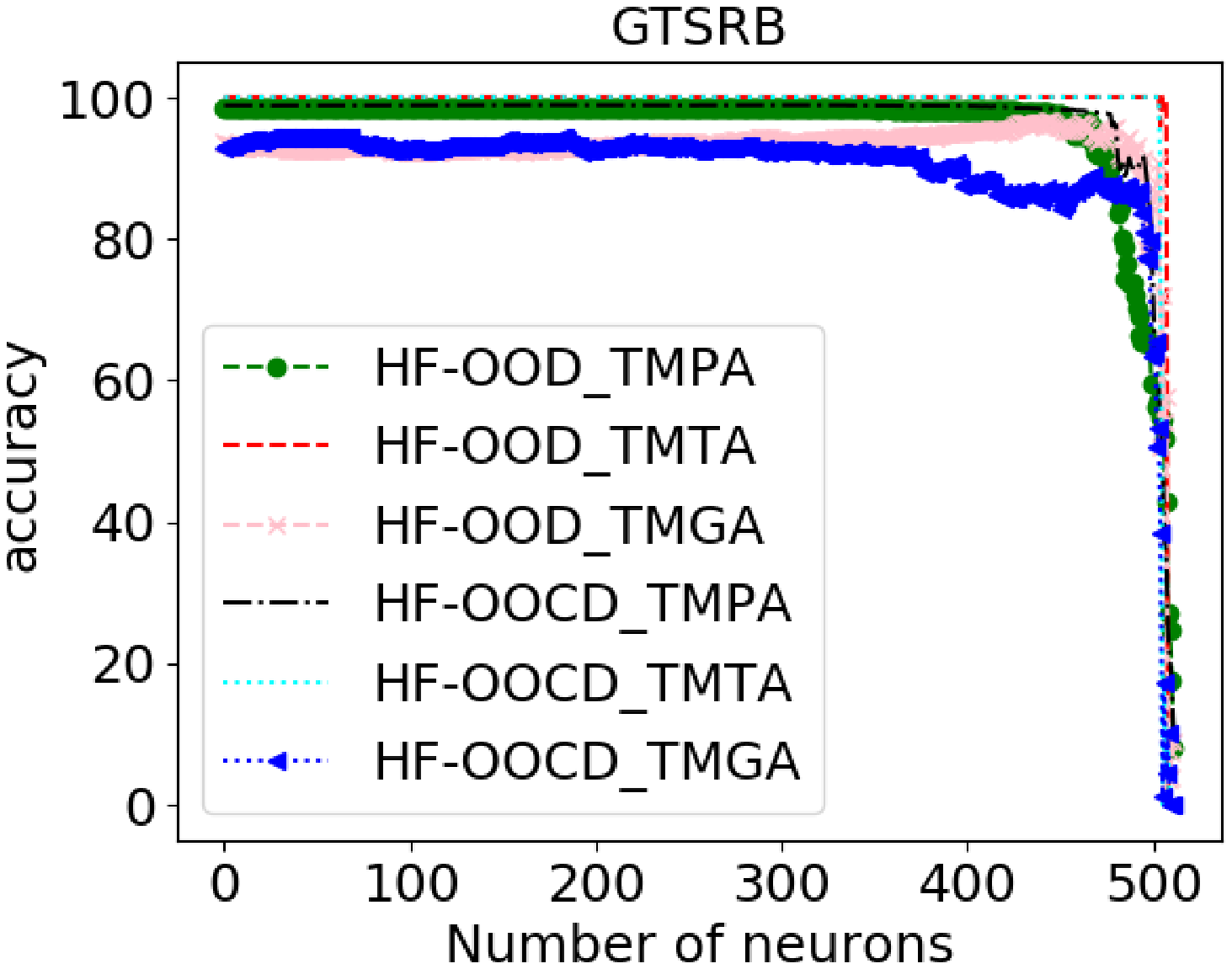}~~\includegraphics[scale=0.35]{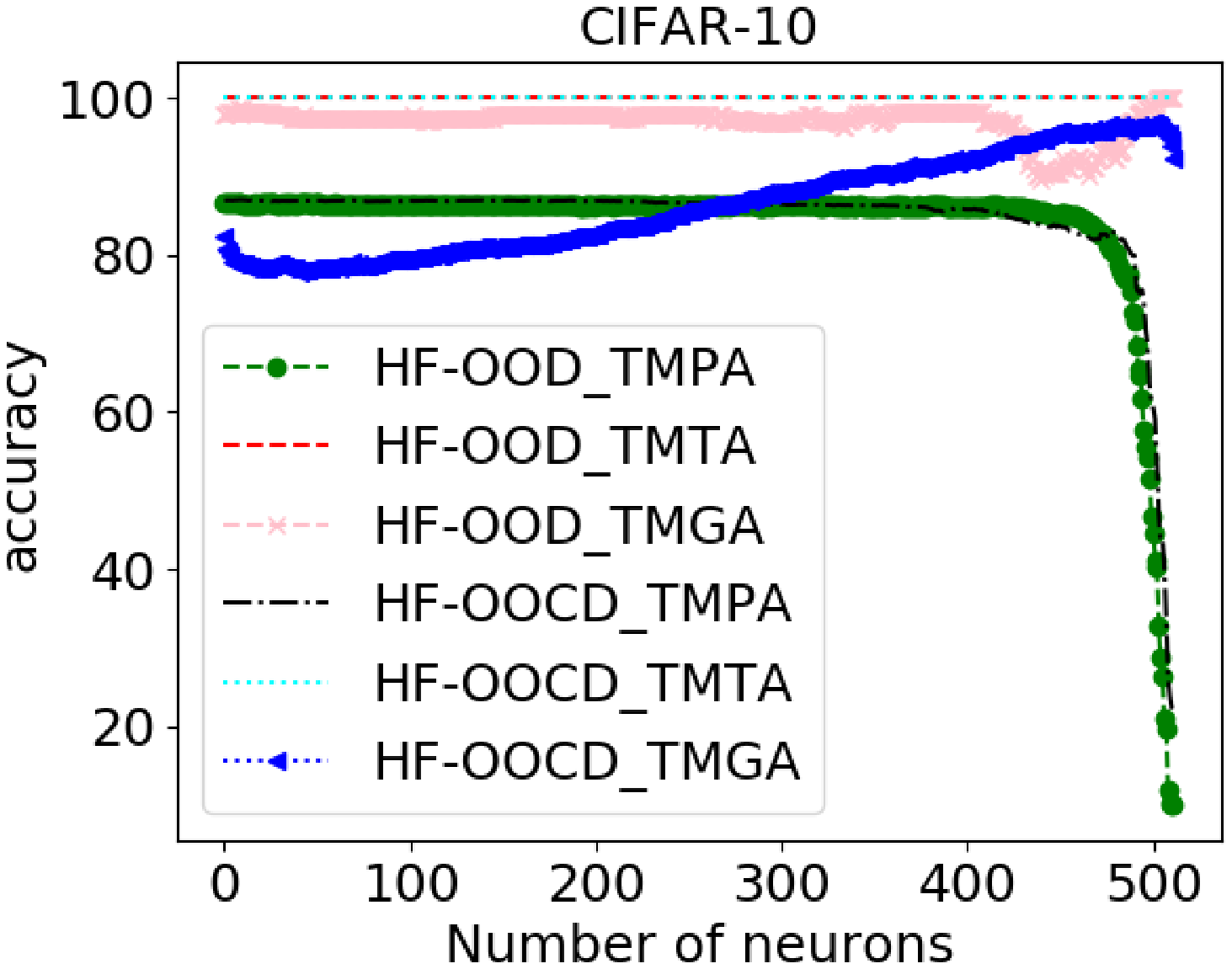}\vspace{3mm}
\caption{Fine Pruning on Preact-Resnet18 HF-OOD and HF-OOCD Trojan models of
GTSRB and CIFAR-10 dataset.\label{fine_pruning}}
\end{figure*}

Table \ref{nc} shows the Neural Cleanse detection performance of
the HF-OOD and HF-OOCD attack models based on the Target class. It
is evident that it has failed to correctly identify the target class
(class with Trojan triggers in it chosen by attacker) all the time
for both GTSRB and CIFAR-10 datasets. In some cases, it has detected
the models as Trojan but based on wrong or non-target classes. That
means the class which is detected as an outlier by NC is not the class
chosen by the attacker. The reverse-engineered patterns (in the supplementary)
show that they failed to detect the Trojan patterns from the target
class. It is clear that the assumption of NC will not be enough to
defend our attack since our trigger is a full-size image than just
a small trigger. Hence, we believe that Neural Cleanse may be behaving
more randomly for our attack scenarios\emph{.}

\subsubsection{Fine pruning }

Fine pruning defense mechanism removes the least activated neurons
based on the pure images the defender has access to. Thus, this mechanism
will prune the neurons which are highly influenced by the Trojan features
and drops the Trojan accuracy of the model. We test the fine Pruning
on the HF-OOD and HF-OOCD Trojan models. The Figure \ref{fine_pruning}
of GTSRB and CIFAR-10 shows \emph{TMPA},\emph{ TMTA}, and \emph{TMGA}
accuracy on the HF-OOD and HF-OOCD Trojan models. It is clear that
the train Trojan data accuracy (\emph{TMTA}) remains intact as the
pruning progresses. It seems to have small drop on the \emph{TMGA}
of GTSRB and CIFAR-10 HF-OOCD and HF-OOD models, however to achieve
it we have to remove most of the neurons from the neural network layer.
It is interesting to note that the CIFAR-10 HF-OOCD \emph{TMGA} improves
as the pruning progresses.
\begin{table*}
\centering{}%
\begin{tabular}{|c|c|c|}
\hline 
\multirow{2}{*}{Attack} & \multicolumn{2}{c|}{\emph{FAR}}\tabularnewline
\cline{2-3} \cline{3-3} 
 & GTSRB & CIFAR-10\tabularnewline
\hline 
input-instance-key attack (Chen\emph{ et al.}\cite{Chen2017}) & 40.0 & 60.0\tabularnewline
\hline 
HF-OOD & 91.0$\pm$11.32 & 99.2$\pm$2.4\tabularnewline
\hline 
HF-OOCD & 100.0 & 92.0\tabularnewline
\hline 
\end{tabular}\vspace{3mm}
\caption{STRIP defense: False Acceptance Ratio (FAR) of a single input-instance-key
attack (Chen\emph{ et al.}\cite{Chen2017}) model, 10 HF-OOD models,
and a single HF-OOCD model on GTSRB and CIFAR-10 datasets. It is evident
from the numbers that our proposed attacks can bypass the STRIP defense
most of the time (90 -100\%).\label{far}}
\end{table*}

\subsubsection{STRIP}

STRIP \cite{Gao_etal_19Strip} is a testing-time defense method, which
classifies an incoming image as Trojan or pure. Given a model and
an input image to test, STRIP will perturb the given input image by
blending it with a set of pure images. If the input image is a Trojan
then the prediction won't change since a fixed trigger does not get
diluted by this process. However, if the input image is not Trojan
then the prediction would differ more widely. Such behaviour is captured
by computing the average class entropy over the perturbed set where
each member of the set is created by blending the input image with
one of the pure images. A low average entropy would indicate the input
image being Trojan.
\begin{figure*}
\begin{centering}
\subfloat[GTSRB\label{strip_gtsrb}]{\begin{centering}
\includegraphics[scale=0.2]{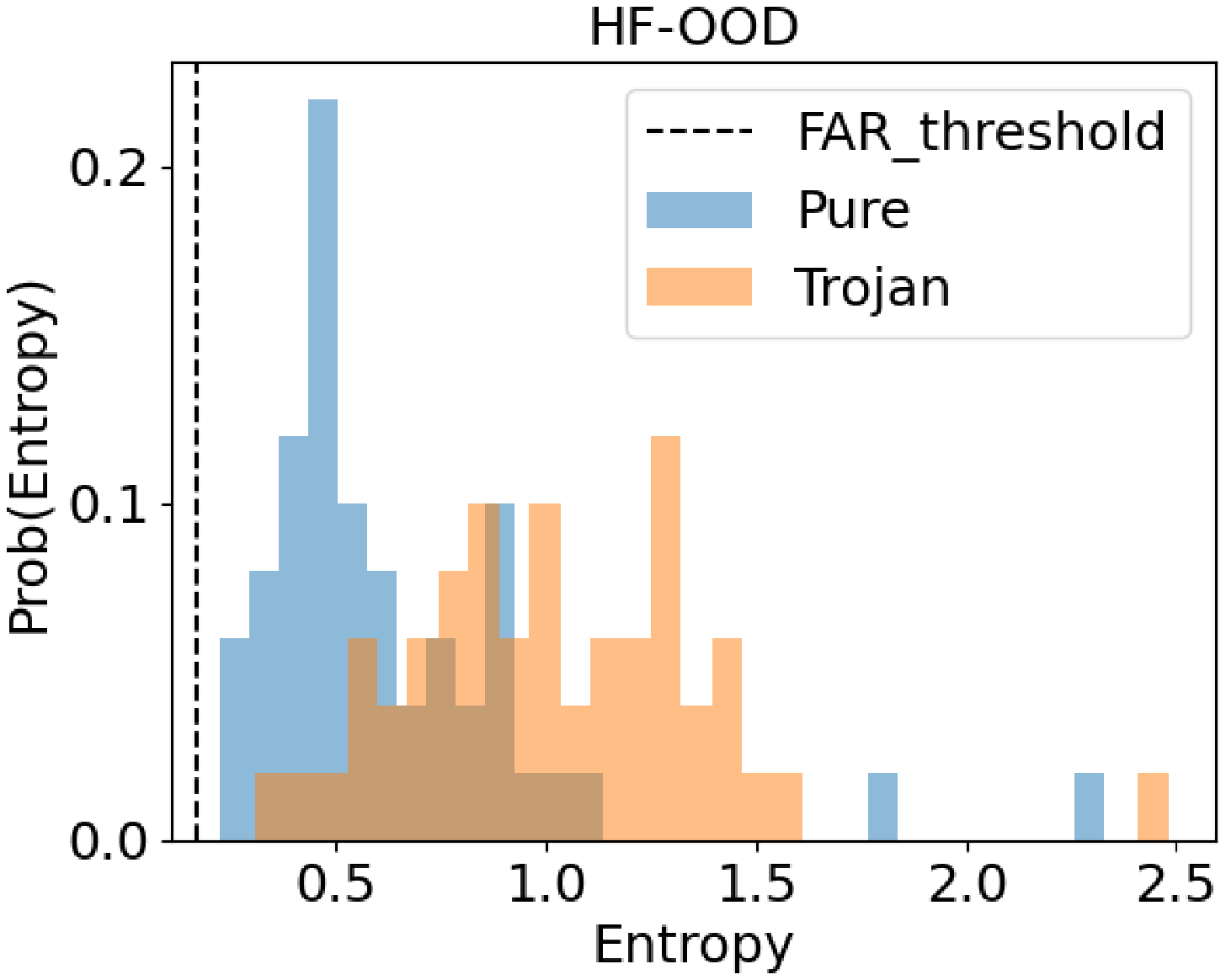}~~\includegraphics[scale=0.2]{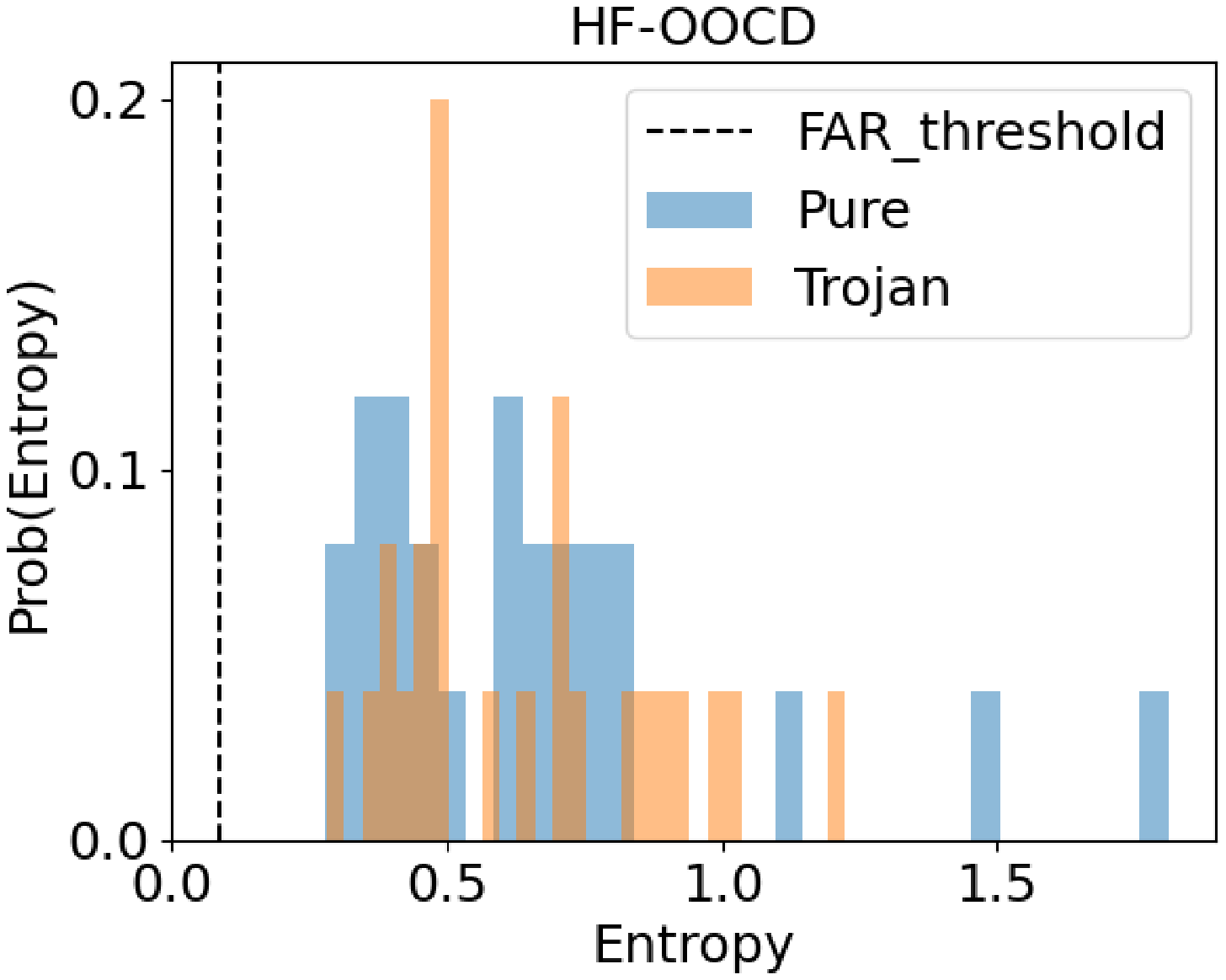}
\par\end{centering}
}\subfloat[CIFAR-10\label{strip_cifar10}]{\begin{centering}
\includegraphics[scale=0.2]{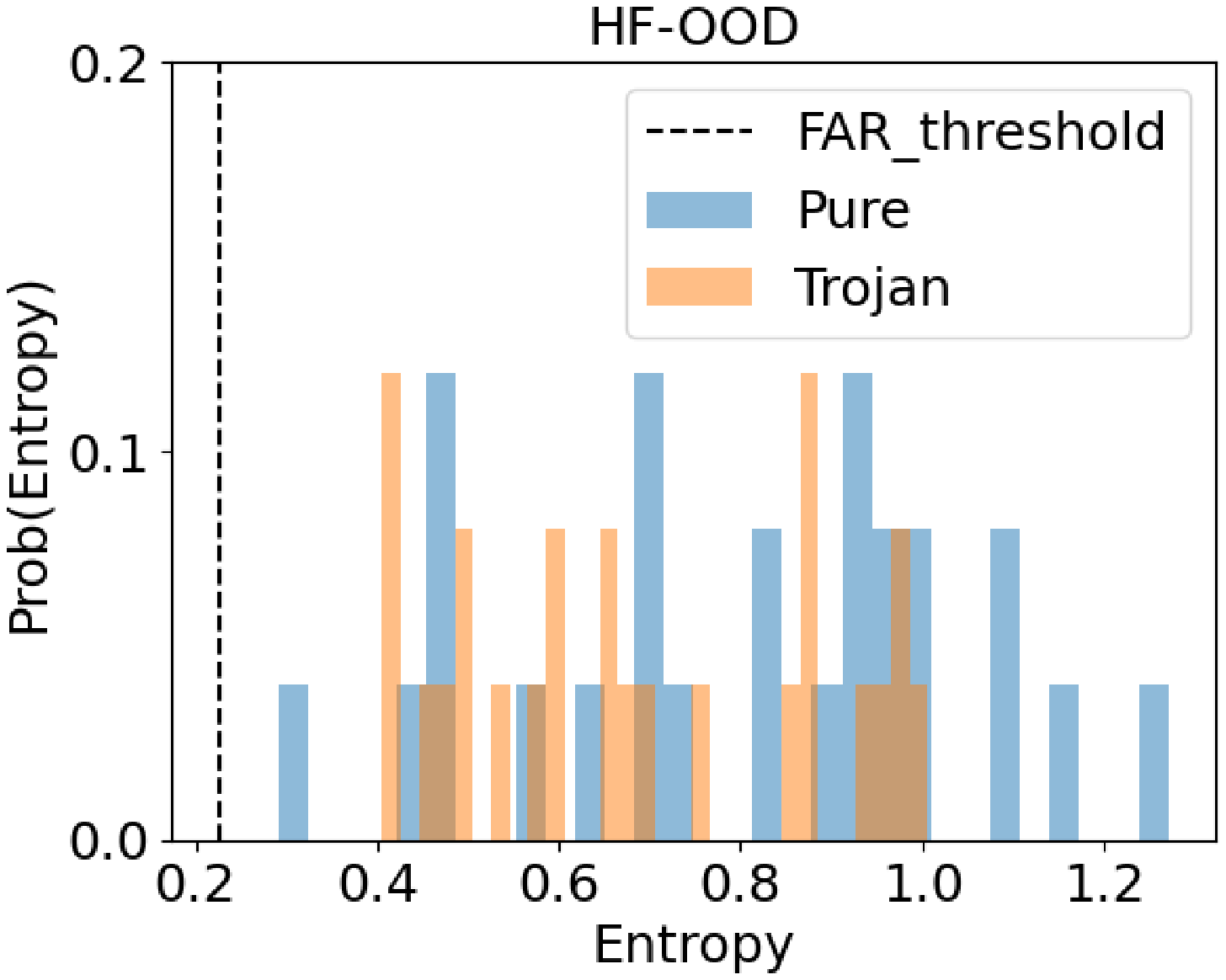}~~\includegraphics[scale=0.2]{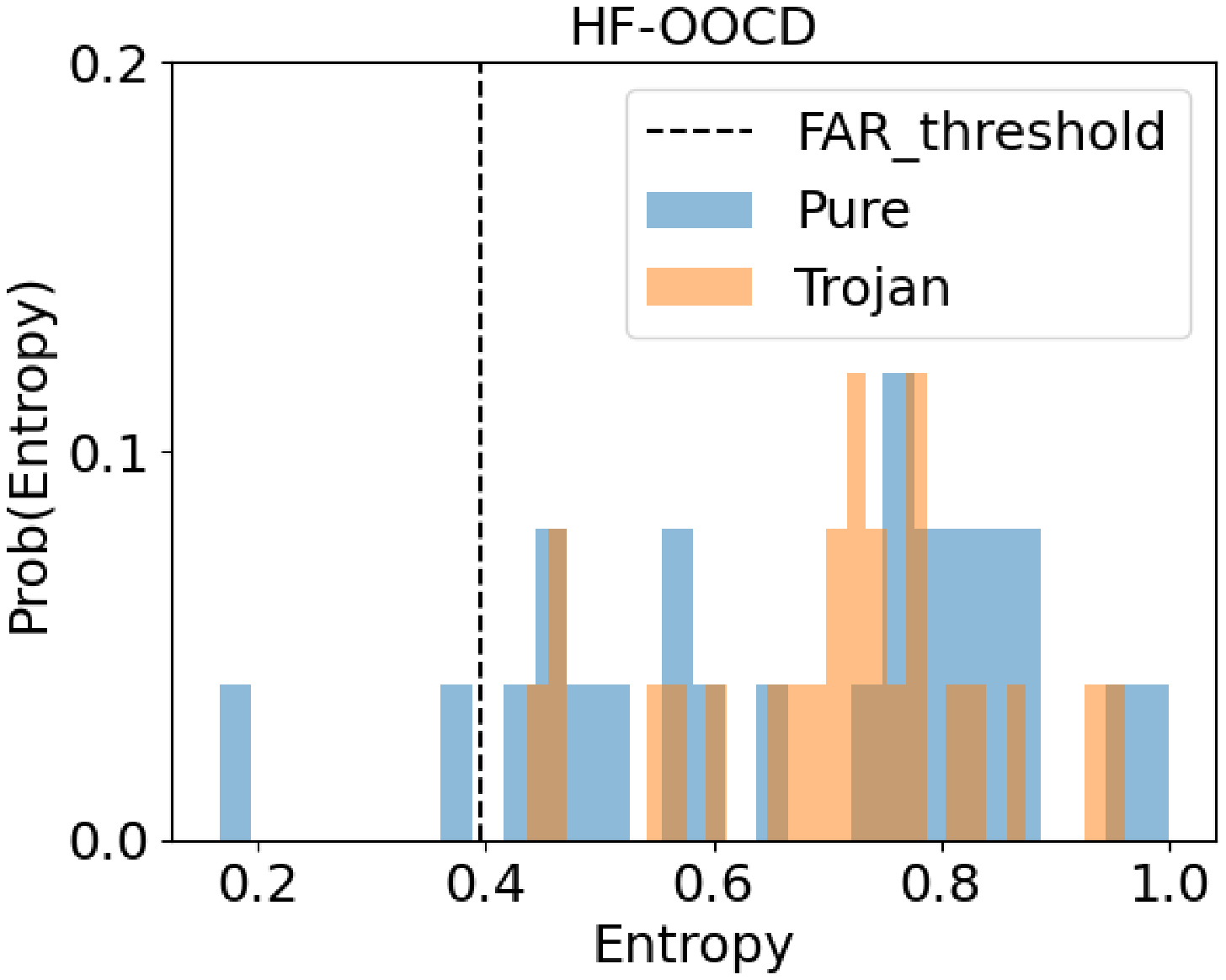}
\par\end{centering}
}
\par\end{centering}
\vspace{3mm}
\caption{STRIP entropy distribution plots on GTSRB (Figure \ref{strip_gtsrb})
and CIFAR-10 (\ref{strip_cifar10}) HF-OOD and HF-OOCD Trojan models.
\label{fig:strip_entrpy_disbtn}}
\end{figure*}

In STRIP, a threshold for the entropy has to be chosen to identify
the given input image as pure. Assume that STRIP defense mechanism
has access to pure images therefore can find the entropy of pure images.
They assume probability distribution of the pure data entropy follows
a normal distribution. The entropy value which belongs to the 1\%
of this normal distribution will be set as the entropy threshold.
Any of the given inputs which has an entropy value above the threshold
will be considered as a pure image. 

We report False Acceptance Ratio (FAR), which is the ratio of the
number of Trojan images accepted as pure images with respect to the
total number of Trojan images in Table \ref{far}. The standard deviation
is added for HF-OOD models since we used 10 HF-OOD Trojan models in
the experiments, however, for the HF-OOCD and input-instance-key we
have a single model for each dataset. The input-instance-key attack
by Chen\emph{ et al.}\cite{Chen2017} perform equally well for Neural
Cleanse and Fine Pruning, however, the performance drops when we use
STRIP.

Figure \ref{fig:strip_entrpy_disbtn} shows the entropy distribution
of a Trojan and pure inputs of a single HF-OOD and HF-OOCD Trojan
model. The plots are similar for other models, as well (please see
supplementary for all of them). It is evident from the figure that
the entropy distributions overlap for both pure and Trojan inputs.
This explains why FAR is high for our attacks. This makes it harder
to find a suitable threshold to separate out the Trojan inputs from
the pure inputs.
\begin{figure*}
\begin{centering}
\subfloat[Accuracy with respect to total number of Trojan samples used for training
the HF-OOD Trojan models for GTSRB (dashed line) and CIFAR-10 (dotted
line) datasets.\label{acc_trn_trojan_samples}]{\includegraphics[scale=0.35]{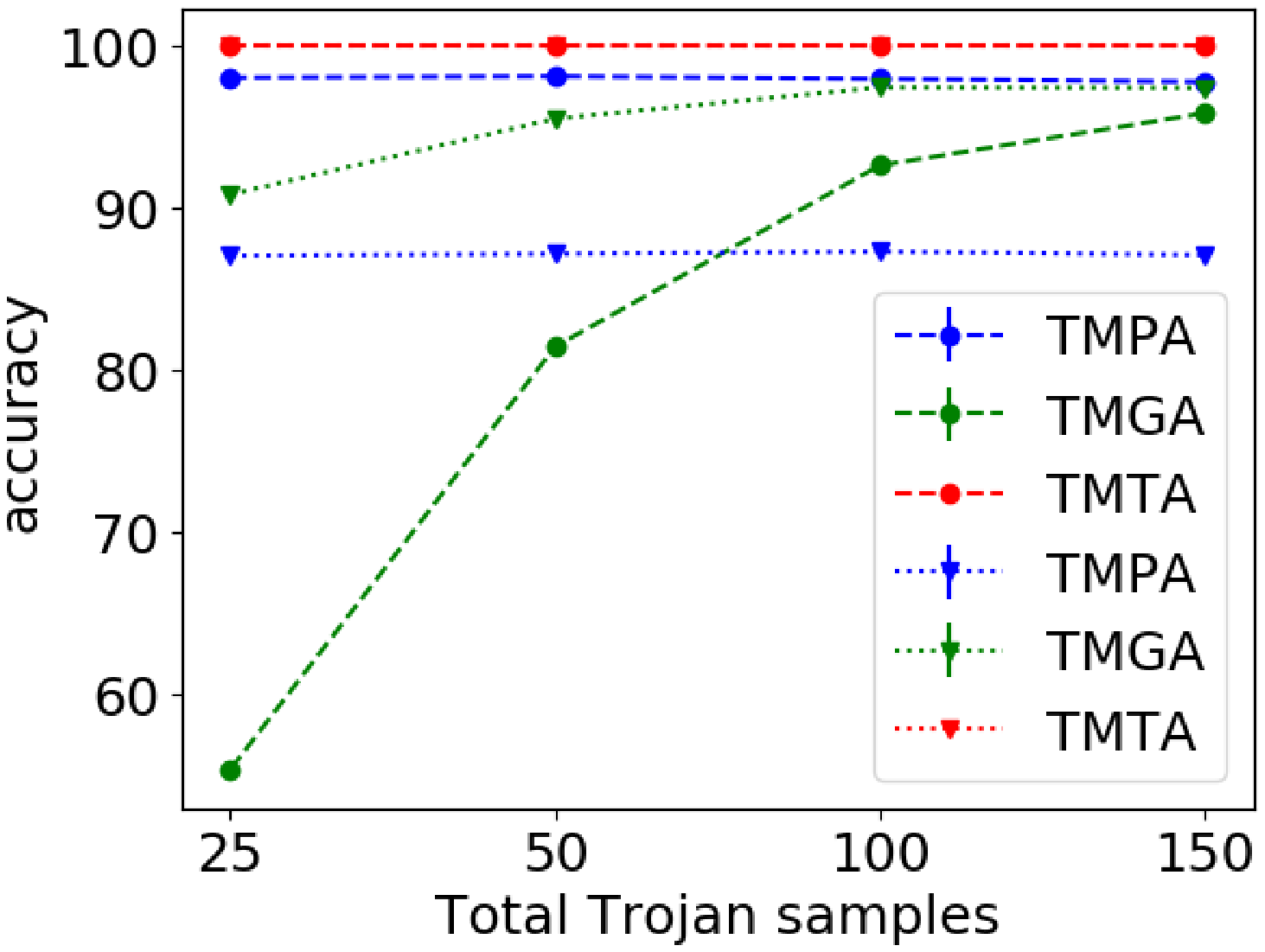}~

}~~~\subfloat[Accuracy with respect to number of Trojan samples in a batch while
training the HF-OOD Trojan models for GTSRB (dashed line) and CIFAR-10
(dotted line) datasets.\label{acc_trn_trojan_btch}]{\includegraphics[scale=0.35]{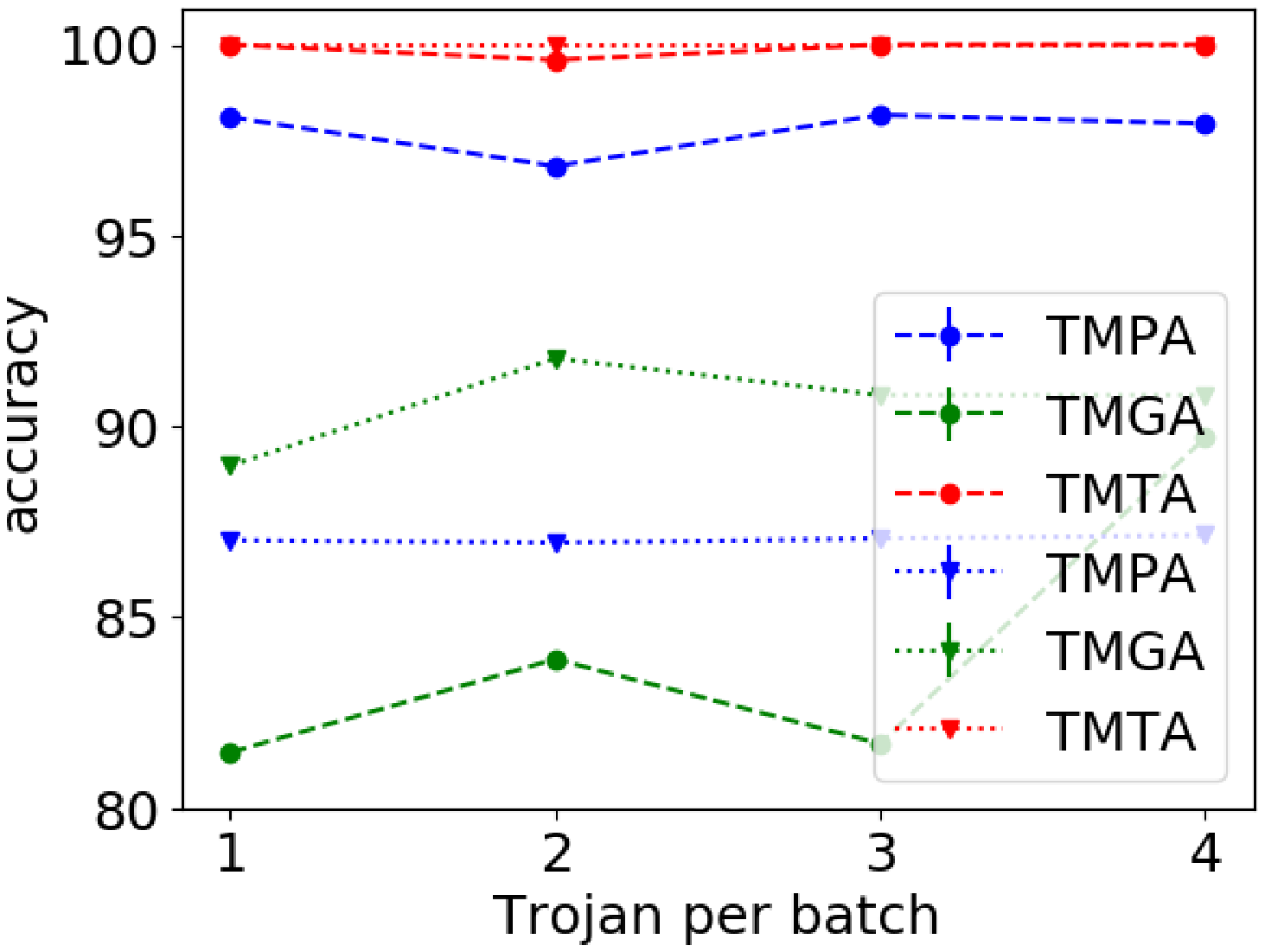}

}\vspace{3mm}
\par\end{centering}
\caption{The average TMPA, TMTA, TMGA accuracies of 10 HF-OOD Trojan models
by varying the total number of Training Trojan data (left) and number
of Trojan per batch (right) for GTSRB (dashed line) and CIFAR-10 dataset
(dotted line).\label{acc_sample_perbatch}}
\end{figure*}
 We have also reported the performance of an input-instance-key attack
method proposed by Chen\emph{ et al.} \cite{Chen2017} and it fails
when we use the STRIP defense mechanism. 

\subsection{Ablation study}

Figure \ref{acc_trn_trojan_samples} shows the change in the average
accuracies of the 10 Trojan models on both GTSRB and CIFAR-10 datasets
with respect to the number of train Trojan samples. It shows that
\emph{TMGA} is the one which gets the most improvement with higher
number of Trojan training images. However, if we use a larger number
of Trojan images it can affect the \emph{TMPA} and can get detected
easily. Figure \ref{acc_trn_trojan_btch} shows the change in the
average accuracies of 10 Trojan models of GTSRB and CIFAR-10 dataset
with respect to the number of Trojan images per batch. For Figure
\ref{acc_trn_trojan_btch} we fix the total number of Trojan images
for training as 25 for CIFAR-10 and 50 for GTSRB. We choose the number
of Trojan training data and Trojan images per batch for our experiments
by analyzing \emph{TMPA}, \emph{TMTA}, and \emph{TMGA} accuracies.

\section{Conclusion}

We introduce a novel and advanced Semantic host-free Trojan attack
in this paper. The pattern learned from the full-size Trojan images
by the backdoored network will be harder to be detected compared to
the existing pattern based attacks. Furthermore, the Trojan images
we introduced are more realistic and remain undetected without a careful
human intervention. We validate through the experiments that the proposed
attack achieves 100\% attack success rate with very few triggers in
model training. Through extensive experiments, we demonstrate the
effectiveness of our semantic host-free attack in defending against
the state-of-the-art defenses. We also shows that the proposed attack
is stealthier compared to a state-of-the-art input-instance-key attack.

\bibliographystyle{plain}
\bibliography{bibs/egbib}

\end{document}